\documentclass[letterpaper, 10 pt, conference]{ieeeconf}  

\IEEEoverridecommandlockouts                              

\usepackage{graphics} 
\usepackage{epsfig} 
\usepackage{mathptmx} 
\usepackage{times} 
\usepackage{amsmath} 
\usepackage{amssymb}  
\usepackage{units}
\usepackage{siunitx}
\usepackage{color}
\usepackage[T1]{fontenc}
\usepackage{booktabs}
\usepackage{multirow}
\usepackage{commath}
\usepackage{amsfonts}
\usepackage{threeparttable} 
\usepackage{tabularx}
\usepackage{cite}
\usepackage{microtype}
\usepackage{hyperref}
\usepackage{multirow}

\let\labelindent\relax
\usepackage{enumitem}

\DeclareMathAlphabet{\mathcal}{OMS}{cmsy}{m}{n} 
\newcolumntype{P}[1]{>{\centering\arraybackslash}p{#1}} 

\usepackage[acronym]{glossaries}
\newacronym{gcn}{GCN}{Graph Convolution Network}
\newacronym{dgcn}{Deep GCN}{Deep Graph Convolution Network}
\newacronym{dptl}{\textit{Descriptellation}}{Deep Learned Constellation Descriptor}
\newacronym{slam}{SLAM}{Simultaneous Localization and Mapping}
\newacronym{sota}{SOTA}{state-of-the-art}
\newacronym{onion}{\textit{Onion}}{Onion Descriptor}
\newacronym{onion hist}{\textit{OHist}}{Onion Histogram Descriptor}
\newacronym{random walk}{\textit{RWalk}}{Random Walk Descriptor}
\newacronym{pointnet}{\textit{PNet}}{PointNet Descriptor}
\newacronym{ransac}{RANSAC}{Random Sample Consensus}
\newacronym{dof}{DoF}{Degrees of Freedom}
\newacronym{qlsm}{QLSM}{Query Local Submap}
\newacronym{mlp}{MLP}{Multilayer Perceptron}
\newacronym{fc}{FC}{fully connected}
\newacronym{gosg}{\textit{GOSG}}{Graph of Semantics Matching - Graph Descriptor}
\newacronym{gosv}{\textit{GOSV}}{Graph of Semantics Matching - Vertex Descriptor}
\newacronym{sgnet}{\textit{SGNet}}{Semantic Graph Networks Descriptor}
\glsdisablehyper

\title{\LARGE \bf Descriptellation: Deep Learned Constellation Descriptors}

\author{
Chunwei Xing$^{\ast,1}$, Xinyu Sun$^{\ast,1}$, Andrei Cramariuc$^1$, Samuel Gull$^1$, Jen Jen Chung$^2$, Cesar Cadena$^1$,\\ Roland Siegwart$^1$, Florian Tschopp$^3$ 

\thanks{$^{\ast}$ Authors contribute to the work equally.}
\thanks{$^1$ Authors are with the Autonomous Systems Lab, ETH Zurich, Switzerland. \tt{\{chxing, xinsun, crandrei, sgull,  cesarc, rsiegwart\}@ethz.ch}.}
\thanks{$^2$ Author is with School of ITEE, The University of Queensland, Australia \tt{jenjen.chung@uq.edu.au}.}
\thanks{$^3$ Author is with Arrival Ltd, United Kingdom, but the work was done while being a member of $^1$. \tt{tschopp@arrival.com}.}
}

\begin{document}

\maketitle
\thispagestyle{empty}
\pagestyle{empty}

\begin{abstract}
Current descriptors for global localization often struggle under vast viewpoint or appearance changes. One possible improvement is the addition of topological information on semantic objects. However, handcrafted topological descriptors are hard to tune and not robust to environmental noise, drastic perspective changes, object occlusion or misdetections. To solve this problem, we formulate a learning-based approach by modelling semantically meaningful object constellations as graphs and using Deep Graph Convolution Networks to map a constellation to a descriptor. We demonstrate the effectiveness of our \gls{dptl} on two real-world datasets. Although \gls{dptl} is trained on randomly generated simulation datasets, it shows good generalization abilities on real-world datasets. \gls{dptl} also outperforms state-of-the-art and handcrafted constellation descriptors for global localization, and is robust to different types of noise. The code is publicly available at \url{https://github.com/ethz-asl/Descriptellation}.

\end{abstract}

\section{Introduction}\label{sec:intro}

A robot's ability to estimate its precise pose in previously visited places is crucial for many applications, such as autonomous driving, trajectory tracking and robotic navigation. One crucial component for localization is a map which records landmarks and to which the robot can match its surroundings. Currently, many maps use appearance-based or object-based descriptors with metric information to represent visited places. Traditional appearance-based descriptors achieve good performance in local mapping, as demonstrated by e.g. ORB-SLAM2~\cite{orbslam} using DBOW~\cite{dbow}. They implement place recognition by selecting a candidate frame that most resembles the query frame, and estimate the pose through keypoints correspondences. However, localization systems based on visual keypoint descriptors, such as SIFT~\cite{SIFT}, SURF~\cite{SURF}, BRIEF~\cite{brief} and ORB~\cite{orb}, suffer from  visual appearance changes due to changing viewpoints and occlusion, leading to candidate mismatches and further localization mistakes. Additionally, the storage requirement increases approximately linearly with the number of frames in the dataset, which limits system scalability on large-scale areas\cite{visualsurvey,capping}.

Constructing maps with object-based descriptors is a proposed solution that uses high-level representations, such as lines~\cite{taubner2020lcd}, cubes~\cite{cubeslam}, quadrics~\cite{quadricslam}, and super quadrics~\cite{tschopp2021superquadric}. 
The works~\cite{bowman2017probabilistic,decentSLAM}, which integrate SLAM with object detection, result in recognizable instance landmarks and inspire the construction of semantic object maps~\cite{decentSLAM}, ideal for robot localization in the presence of changing views and occlusion.
Some proposed descriptors~\cite{taubner2020lcd, 7353986,liu2019global} use the semantic label directly as a descriptor. However, just the semantic label is often not distinct enough, especially when many objects of the same category are present. It is time-consuming to pair target and query objects with such aliasing descriptors. In contrast, a semantic constellation with structural cues, consisting of an object and its neighbors, is likely to be unique for a given scene~\cite{xview,liu2019global}. We map the structural and semantic information of an object constellation to a descriptor for global localization. 
 
\begin{figure}[t]
      \centering
      \includegraphics[width=\columnwidth]{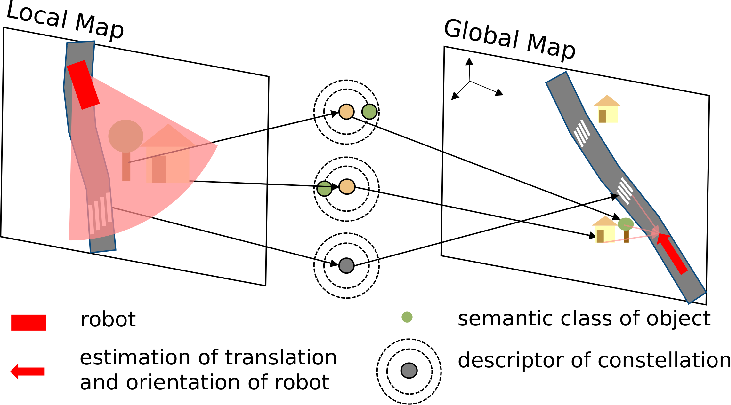}
      \caption{Global localization based on semantic constellation descriptors. The global and local semantic object maps are described as object constellations. For each object in the global and local map, we generate a graph by specifying an edge between every pair of neighboring nodes according to the Euclidean distance. The graph is used to generate robust descriptors using \glspl{gcn}.}
      \label{fig:motiv}
\end{figure}

\glspl{gcn} show great potential in solving complex constellation classification problems~\cite{ssclassgcn2016,dynamicgraph}. In this work, we introduce a learning-based semantic constellation descriptor combining both geometric and semantic information. An overview of our method is shown in Figure~\ref{fig:motiv}. We extract object constellations from object-centric maps and construct graphs from them. We incorporate \glspl{gcn} for representation learning from these graphs with embedded semantic information in each vertex to obtain descriptors for global localization. Our work can be divided into three main contributions:
\begin{itemize}[leftmargin=*]
\item A graph-based learned descriptor of fused geometric and semantic information extracted from object constellations.
\item A pipeline to train our deep-learned descriptor on simulation data and to deploy it to real-world applications.
\item Comparisons on real-world datasets against existing baselines showing the feasibility and effectiveness of \gls{dptl} in terms of localization accuracy.
\end{itemize}

\begin{figure*}[htbp]
    \centering
    \includegraphics[width=0.75\textwidth]{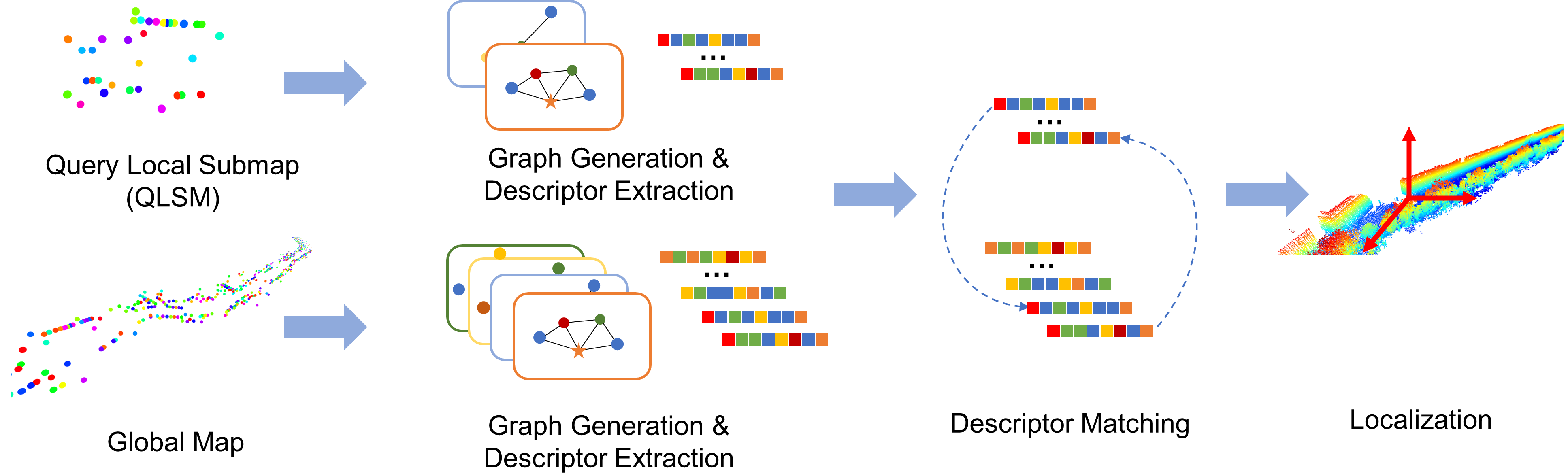}
    \caption{Overview of the global localization system based on \gls{dptl}. For each object in the \gls{qlsm} and the global map, a graph is built from the constellation of its neighboring objects, and its descriptor is extracted from the graph. By matching the descriptors in the \gls{qlsm} with the database (descriptors extracted for each objects in the global map), candidate matching pairs can be found. By performing geometric verification on these matching pairs using RANSAC, a transformation matrix from the \gls{qlsm} frame to the world frame can be obtained.}
    \label{fig:overview}
\end{figure*}

\section{Related Work}\label{sec:relate work}

\subsection{Descriptors for Global Localization}\label{sec:desc refs}
To achieve global localization, many descriptor generation methods exist, which typically trade off between distinctiveness and robustness. Distinctiveness reflects whether each landmark in a global map has an unique representation, which reduces the matching time when looking for pairwise correspondences. Robustness means that the same landmark's descriptor should be consistent between maps and invariant to perspective changes, occlusion, and the passage of time. In terms of whether high-level structural information is included, descriptors can be categorized as \emph{local} or \emph{contextual}. 

Local descriptors extract visual features around keypoints~\cite{SHOT,local_desc}, or geometric features of segments~\cite{segmatch,segmap,semsegmap} and show high reliability in place recognition. Contextual descriptors such as VLAD~\cite{VLAD}, PointNetVLAD~\cite{pointnetvlad}, or 3D Gestalt~\cite{6630945} aggregate features from neighborhoods and are robust against noise. However, these descriptors are not rotation-invariant and perform poorly under viewpoint changes. 

Recently, contextual descriptors from semantic graphs~\cite{xview,semantic_graph} have been used for place recognition, achieving rotation-invariant performance in several datasets. Semantic objects form the nodes of the graph as 3D points, which can be connected by edges based on \textit{e.g.} proximity. Such graphs contain both geometric and semantic information, and thus can be used to extract fused features that are more robust to the aforementioned changes~\cite{semantic_graph}. \glspl{random walk}~\cite{xview, liu2019global} achieve good performance on graph matching and localization. However, \glspl{random walk} implicitly utilize the topological information of the graph instead of the explicit geometric information, and performance varies with the walk step count and graph complexity.

\subsection{Deep Learned Graph Descriptors}\label{sec:learning desc refs}
Inspired by the success of \glspl{gcn}, graph embeddings learned by graph similarity networks are used to match two graphs built from semantic maps~\cite{graph_review,semantic_graph}. The graph networks concatenate the spatial and semantic features extracted by two edge convolution networks directly as the graph embedding. However, the geometric and semantic information could be combined as inputs and features extracted concurrently. 

Graph representation learning methods can learn node-level and graph-level embeddings. Node-level embedding methods such as DeepWalk~\cite{deepwalk}, node2vec~\cite{node2vec}, and Bag-of-Vectors~\cite{BOV} extract local graph structural information and project it onto lower dimensions. However, these methods lose the global structural information when extracting features. In this paper, we propose a graph-level embedding as the descriptor of an object by aggregating both node and edge features. \glspl{gcn} extract structural information of neighboring nodes using spatial convolution operations.

In this paper, we use \glspl{dgcn}~\cite{li2019deepgcns} to extract node features and use the global attention layer~\cite{global_attention} to aggregate all the features extracted by \glspl{dgcn} into a fixed-length graph-level embedding which can be used as a global descriptor.

\section{Method}\label{sec:method}

In this section, we present our localization system based on \gls{dptl}. It leverages graph extraction from semantic object-centric maps (\ref{sec:extract_cstl}) and graph matching using graph descriptors (\ref{sec:feature_learn}) for global localization (\ref{sec:global_local}). Figure~\ref{fig:overview} illustrates the overview of our system, focusing on the graph representation and matching. 

Our goal is to localize a robot by matching the locally built map, called \textit{\glsfirst{qlsm}}, with a previously recorded \textit{global map}.
In both maps, each node (\textit{i.e.} object) is assigned a learned descriptor by formulating its surrounding sub-constellation as a graph and extracting that graph's features. Localization is achieved by matching descriptors of the \gls{qlsm} to descriptors of the global map.

\subsection{Object Constellation Extraction and Graph Generation}\label{sec:extract_cstl}
 
The global and local semantic object maps are modeled as object constellations. For each object in the global and local map, we generate a graph by specifying an edge between every pair of neighboring objects (or nodes) according to the Euclidean distance within a predefined threshold, including the self-loop of the object itself. The threshold is defined by the average distance between two nodes in the global map.

\subsection{Network Architecture}\label{sec:feature_learn}
\begin{figure*}[htbp]
    \centering
    \includegraphics[width=\textwidth]{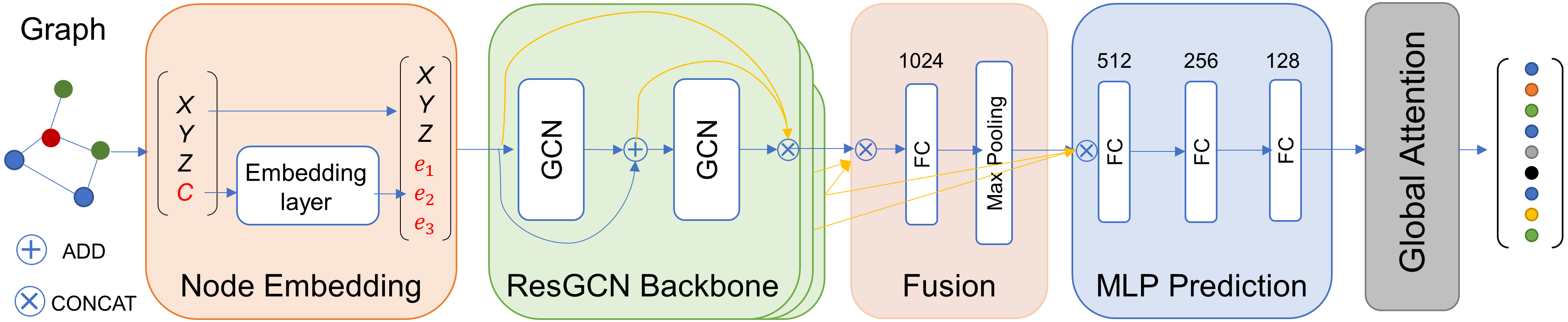}
    \caption{Network architecture: Given a graph constructed according to Section~\ref{sec:extract_cstl}, the network integrates the spatial and semantic information of the edges and nodes to generate a graph descriptor using an extension of the \gls{dgcn} architecture~\cite{li2019deepgcns}. The numbers above the fully connected (FC) blocks represent output sizes.}
    \label{fig:sgbpr}
    \vspace{-0.5cm}
\end{figure*}

As shown in Figure~\ref{fig:sgbpr}, our network is adapted from \gls{dgcn}~\cite{li2019deepgcns}  consisting of the ResGCN backbone, a fusion block and a \gls{mlp} prediction block. We modify the original network by adding an input embedding layer and an output global attention layer.
The input of each node in the graph contains the 3D coordinates of its centroid $(x, y, z)$ and an class label $C$. A standard one-hot encoding of the class label would considerably increase the size of the network. The ablation study on the input encoding methods refers to Section~\ref{sec: training ablation}. We therefore preprocess the label by learning a 3-dimensional embedding $e$. To extract features from node embeddings and edges, we use a 14-layer ResGCN as a backbone, and a Max-Relative convolution layer. The fusion block consists of a fully connected layer with max-pooling.

In order to aggregate the embeddings of all the nodes into a graph-level fixed-length embedding, an attention layer~\cite{global_attention} is used. It is a gated graph sequence neural network given by \begin{equation}
    \mathbf{r}_i = \sum_{n=1}^{N_i} \mathrm{softmax} \left(
        h_{\mathrm{gate}} ( \mathbf{x}_n ) \right) \odot
        h_{\mathbf{\theta}} ( \mathbf{x}_n ),
    \label{eq:attenion}
\end{equation}
where $h_{\mathrm{gate}}$ and $h_{\mathbf{\theta}}$ are neural networks, $\mathbf{x}_n$ denotes the node features for node $n$, $N_i$ is the number of nodes for batch index $i$, $\mathbf{r}_i$ denotes the batch-wise graph-level output, and $\odot$ denotes element-wise multiplication.
We utilize the batch triplet loss~\cite{batchtriplet2017} to optimize the parameters during training. 

\subsection{Global Localization}
\label{sec:global_local}
To account for outliers and estimate a 3D transformation between the \gls{qlsm} and the \textit{global map} we use \gls{ransac}~\cite{fischler1981random}. For each object descriptor in the \gls{qlsm}, we find the corresponding $K$ candidates with the closest descriptors in $L2$ norm space in the \textit{global map}. \gls{ransac} then finds a 3D geometric registration between queries and candidates given the predefined minimum inlier threshold $t_{ransac}$. Addressing applications with observable gravity direction (e.g. ground vehicles or visual-inertial mapping) we assume a gravity-aligned local submap frame. Furthermore, the translation along $z$ is negligible, and thus we only optimize for $(x, y, yaw)$. Nevertheless, the full 6 \gls{dof} poses are still recoverable using the same method if the application requires it.

\subsection{Simulated Training Dataset Generation}
\label{sec:traindata}
The training dataset is created in a simple simulation environment by stochastically generating a series of object-centric maps with different objects. A map is populated by superimposing randomly generated set of objects, using various geometric patterns: lines, circles, and multivariate normal distributions. The number of such patterns, the number of nodes in each pattern and the semantic labels of objects are sampled from a discrete uniform distribution. The relative coordinates of the patterns are sampled from a continuous uniform distribution.

Afterwards for each object, its constellation is first created as the anchor, by including the surrounding objects and with the selected object as the center node. By transforming the anchor constellation using randomly sampled viewpoints, positive samples are obtained. A graph is then constructed from each constellation using the method illustrated in Section~\ref{sec:extract_cstl}. For the training and validation data, we generate 1000 anchor constellations and 9 positive constellations for each anchor using this strategy.

We simulate inconsistent observation of visited places, that happens in real-world scenarios, by adding noise when sampling the training data. This data augmentation additionally helps to prevent overfitting. We define $\mathbf{p}_{i}$ as the 3D position of object $i \in \mathbb{C}$ with respect to the origin of the constellation $\mathbb{C}$, and $p_{i, \xi}$ is the position along the $\xi$-axis$\,\in \{x, y, z\}$. $\mathcal{N}$ and  $\mathcal{U}$ denote normal distribution and uniform distribution respectively. We use the following augmentation methods:
\begin{itemize}[leftmargin=*]
    \item \textit{Trans}: Translational noise is added to simulate the localization errors from upstream tasks, \textit{e.g.} object mapping. This noise is added to each object in the constellation by 
    \begin{equation} \tilde{p}_{i, \xi} = p_{i, \xi}+\left\Vert\mathbf{p}_{i}\right\Vert\cdot e_{trans},
    \end{equation}
    where $\mathbf{\tilde{p}}_{i}$ defines a noisy position, and $e_{trans} \sim \mathcal{N}(0, 0.25)$.
    \item \textit{Orient}: To simulate perspective changes, orientational noise is added to the whole constellation, \textit{i.e.}, the constellation is rotated around its origin by $r \sim \mathcal{U}(-\ang{180}, \ang{180})$.
    \item \textit{Dropout}: This noise is used to simulate the \textit{false negative} ratio of an upstream task, \textit{e.g.} object detection. We remove an object with a probability $e_{dropout} = 0.1$.
    \item \textit{FP}: This noise is implemented in the same context as \textit{Dropout} noise, but instead represents the \textit{false positive} ratio. We add a random object with a probability $e_{FP} = 0.1$.
    \item \textit{Misclass}: Misclassification noise represents the upstream classification task error rate. We change the semantic label of each object $i\in\mathbb{C}$ with a probability 
    \begin{equation}
    e_{misclass} = \left\Vert\mathbf{p}_{i}\right\Vert / \max_{j\in\mathbb{C}}\left\Vert\mathbf{p}_{j}\right\Vert \cdot \alpha_{misclass},
    \end{equation}
    where $\alpha_{misclass} = 0.1$.
    \item \textit{Crop}: We crop the constellation by removing objects along the $\xi$-axis if
    \begin{equation}p_{i,\xi} > \max_{j\in\mathbb{C}}\left\Vert\mathbf{p}_{j}\right\Vert \cdot (1 - e_{crop}),
    \end{equation}
    where $e_{crop} \sim \mathcal{U}(0, 0.3)$.
    \item \textit{Scale}: The scale noise is applied to the whole constellation. The coordinates of each object are scaled according to the same factor $s \sim \mathcal{U}(0.85, 1.25)$.
\end{itemize}

\subsection{Implementation}\label{sec:train detail}
Our entire model is trained from scratch in an end-to-end fashion using the Adam optimizer~\cite{kingma2014adam} for 100 epochs, and the exponential decay rates $\beta_1$ and $\beta_2$  for the moment estimates are 0.9 and 0.999 respectively. The initial learning rate is $1\times 10^{-3}$, and the decay rate is $0.7$. The training took approximately 3 hours on a NVIDIA GeForce GTX 1060 Ti GPU and 6 Intel i7-8750H CPU. 

We use the \textit{topK} ratio as a target objective to stop the training and select the best model. More specifically, at the end of each epoch for all constellation embeddings in the dataset, the $K$ closest embeddings among all the other constellations are selected. The \textit{topK} ratio is defined as the ratio of samples where at least one of the $K$ selected embeddings has a matching label. The \textit{topK} ratio strongly relates to the localization efficiency since the effectiveness of \gls{ransac} highly depends on the number of false candidates it has to sort through.

\section{Experiments}\label{sec:exps}

\subsection{Datasets}\label{sec:dataset}

To test the localization performance of \gls{dptl} on real world scenarios, we use two outdoor point cloud datasets: Paris-Rue-Lille~\cite{lille} and IQmulus~\cite{cassette}. These datasets contain fine-grained point cloud data captured in a driving scenario, with semantic and instance annotations. 

\begin{table}[htbp]
    \centering
    \caption{Dataset statistics, before and after preprocessing.}
    \label{tab:data stat}
    \begin{tabular}{cc|cc|cc}
    \toprule
         \multirow{2}{*}{\textbf{Trajectory}} & \multirow{2}{*}{\textbf{Length}} &  \multicolumn{2}{c|}{\textbf{Original}}  & \multicolumn{2}{c}{\textbf{Preprocessed}} \\ 
          & & Objects & Classes & Objects & Classes \\ \midrule
         \texttt{Lille1} & \SI{1150}{\metre} & 1349 & 39 & 795 & 25 \\
         \texttt{Lille2} & \SI{340}{\metre} & 501 & 29 & 274 & 19\\
         \texttt{Paris} & \SI{450}{\metre} & 629 & 41 & 306 & 23 \\ 
         \texttt{IQmulus} & \SI{200}{\metre} & 414 & 22 & 152 & 13 \\
    \bottomrule
    \end{tabular}
    
\end{table}

The Paris-Rue-Lille~\cite{lille} is an urban dataset, containing 3 trajectories: \texttt{Lille1}, \texttt{Lille2}, and \texttt{Paris}. The \texttt{IQmulus} database~\cite{cassette} is scanned from a dense urban environment. The dataset statistics are summarized in Table~\ref{tab:data stat}. 

Dynamic objects, such as pedestrians and cars, are removed in pre-processing since we do not consider them relevant for localization. We want to focus on objects we will be able to redetect when revisiting the place. Similarly, unknown object categories are also removed. In addition to removing semantically dynamic objects, we also remove the ``bollard" objects from the dataset, because we find the distribution of these objects is very dense across the whole dataset and they appear in very repetitive patterns. While \gls{dptl} still outperforms the other methods when using ``bollards", their addition decreases the performances of all baselines significantly due to heavy self-similarity. We represent the remaining objects using a 3D point located in the object's centroid.

\subsection{Experiment Setup}\label{sec:exp setup}

\subsubsection{Object-centric Maps Generation}\label{sec:generate map}

We generate object-centric maps from the point cloud datasets. The densely segmented global point cloud map is downsampled into voxels, each including 3D coordinates, an instance ID, and a semantic class label. The centroid of an object is computed as the mean coordinates of voxels with the same instance ID. The global map is then composed of all these object centroids. Each \textit{semantic object} is then represented by the 3D centroid coordinates, the instance ID and the semantic class label. To extract descriptors for an object, we retrieve a local constellation for each \textit{semantic object} in the global map. The local constellation is defined as a union of the object itself and its neighboring objects within a visual range threshold.

\subsubsection{Data Collection}\label{sec:data collect}
When collecting the test data, we randomly sample query positions and orientations on the predefined sampling trajectory in the object-centric map. For each query position, we take the current visible static objects from the query position within a visual range threshold to construct the \textit{\gls{qlsm}}, and we remove occluded objects~\cite{hidden_point}. We set the default visual range threshold to \unit[30]{m}. The coordinates of each object in the map are transformed into local coordinates with a random yaw angle to simulate the observation at an unknown orientation. Object constellation extraction and graph generation then follow Section~\ref{sec:extract_cstl}. 


\subsubsection{Testing Scenarios}\label{sec:test scenes}
We propose 3 experiment cases to test the global localization performance of \gls{dptl}.
\begin{itemize}[leftmargin=*]
    \item \textit{Self-localization}: In this experiment we use unpolluted data in the \gls{qlsm}. This scenario represents testing with a perfect upstream object mapping system.
    \item \textit{Fewer Objects}: To simulate the realistic scenario where objects are deficient or they could only be reached in a limited range, we reduce the visual range threshold to \unit[20]{m} from the centric object when building the \gls{qlsm}.
    \item \textit{Added Noise}: Since the trajectories of the dataset were collected only once, to represent a more realistic localization scenario, we introduce different kinds of noise, including \textit{Trans}, \textit{Dropout}, \textit{Misclass}, and \textit{Scale}. The noise is implemented in the same way as described in Section~\ref{sec:traindata}. For the \textit{Trans} and \textit{Scale} noises, we choose the error rate $e_{trans} \sim \mathcal{U}(0, 0.1)$ and scale ratio $s \sim \mathcal{U}(0.9, 1.1)$. For the \textit{Dropout} and \textit{Misclass} noises, considering the accuracy of state-of-the-art object detection and image classification models~\cite{vora2020pointpainting}, we choose $e_{dropout} = 0.1$ and $\alpha_{misclass} = 0.2$.
\end{itemize}

\subsection{Baselines}\label{sec:baseline}
We compare \gls{dptl} to five handcrafted descriptors: \acrfull{onion}, \acrfull{onion hist}~\cite{onionhist}, \acrfull{random walk}~\cite{xview}, \acrfull{gosg}~\cite{gosmatch}, and \acrfull{gosv}~\cite{gosmatch}.  Additionally, we also compare to two learned descriptors: \acrfull{pointnet}~\cite{qi2017pointnet}, and \gls{sgnet}~\cite{semantic_graph}.

\begin{figure}[t]
    \centering
    \includegraphics[width=0.5\textwidth]{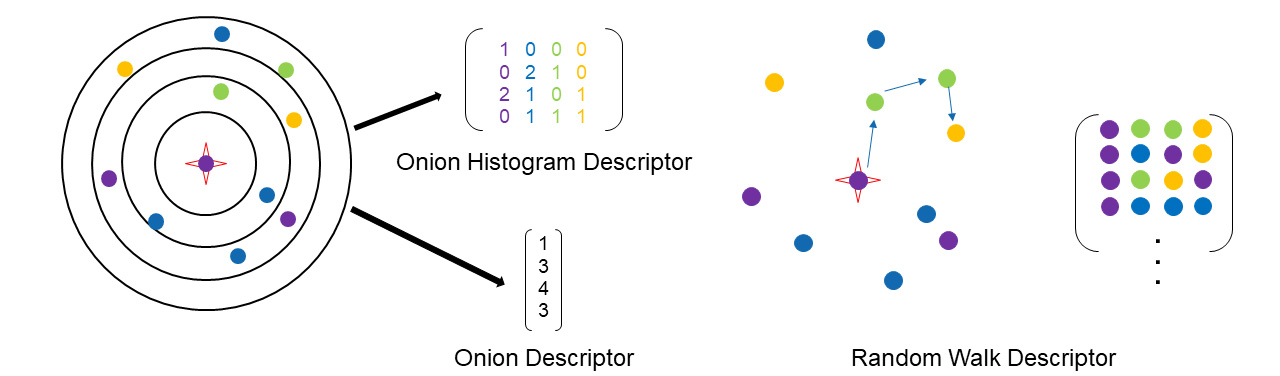}
    \caption{Handcrafted baseline descriptors. The onion and onion histogram descriptor example shows the case when the number of shells $n_s = 4$ and number of classes $n_c = 4$. The random walk descriptor example shows the case when the walk length $l_w = 4$.}
    \label{fig:baseline}
\end{figure}

Illustrations of how the handcrafted descriptors work are shown in Figure~\ref{fig:baseline}. For \gls{onion} and \gls{onion hist}, objects are split into $n_s$ shells with equal spacing, $d_s$, between two neighboring shells. \gls{onion} counts the number of objects in each shell as the descriptor, and \gls{onion hist} creates a histogram of objects' semantic classes instead. \gls{gosv} shares a common format with \gls{onion hist}, but instead of counting semantic objects within a certain distance, it counts all the vertices in the whole \gls{qlsm}~\cite{gosmatch}. Similarly, \gls{gosg} counts all the edges and bins them by edge type and distance~\cite{gosmatch}. For \gls{random walk}, a number of $n_w$ random walks of length $l_w$ are sampled starting from the query object. The semantic class labels of the visited objects are then stored in a descriptor matrix of shape $n_w \times l_w$. 

We tune the parameters of the handcrafted baseline descriptors on simulation data. For \gls{onion} and \gls{onion hist}, we chose $n_s = 3$ and $d_s = \SI{10}{\metre}$. For \gls{random walk}, we chose the walk length $l_w = 4$ and the number of walks $n_w = 30$. We keep the parameters the same as in the original paper for \gls{gosg} and \gls{gosv}~\cite{gosmatch}. For \gls{pointnet}, we adapt the original PointNet++~\cite{qi2017pointnet} by removing the soft-max layer at the end of the network and train it using the same data, loss, and metrics as \gls{dptl}. For \gls{sgnet}, we use the best pre-trained model for evaluation by remapping the semantics. 

\subsection{Evaluation Metric}\label{sec:eva}
We evaluate the global localization performance by the success rate defined on the translation errors. In detail, the success rate $\eta$ is defined as the ratio of localized samples whose translation errors are less than $\unit[1]{m}$.

\begin{table}[ht!]
\setlength\tabcolsep{4pt}
\renewcommand{\arraystretch}{1.2} 
\small
\caption{Global localization performance on the listed datasets. The best descriptor in each scenario is highlighted in bold text. }
\label{tab:test res}
\begin{center}

\begin{tabular}{@{}l|c|ccc@{}}

\hline
\hline
 \parbox[t]{2mm}{\multirow{2}{*}{\rotatebox[origin=c]{90}{\texttt{Traj}}}}
 & Metrics &\multicolumn{3}{c}{Translation Success Rate$\left(\eta\right)\left[\%\right]$} \\ 
 & Scenarios & Self-localization & Fewer Objects & Added Noise \\
 
\hline
\hline
 \parbox[t]{4mm}{\multirow{7}{*}{\rotatebox[origin=c]{90}{\texttt{Lille1}}}} 
 &\gls{onion}       & 4.00 ± 0.31         & 1.48 ± 0.57   & 0.00 ± 0.00   \\
 &\gls{onion hist}~\cite{onionhist}  & 96.80 ± 0.80        & 90.88 ± 1.13  & 32.36 ± 0.91  \\
&\gls{random walk}~\cite{xview} & 93.24 ± 1.94        & 82.64 ± 0.81  & 40.92 ± 1.51  \\
 &\gls{gosg}~\cite{gosmatch}        & 70.08 ± 0.93        & 58.08 ± 0.57  & 7.80 ± 1.12  \\
 &\gls{gosv}~\cite{gosmatch}         & 91.72 ± 0.60        & 79.48 ± 0.16  & 30.44 ± 1.49  \\
 &\gls{pointnet}~\cite{qi2017pointnet}    & 6.92 ± 0.48         & 4.32 ± 0.68   & 1.08 ± 0.30   \\
 &\gls{sgnet}~\cite{semantic_graph}       & 73.52 ± 1.22        & 49.32 ± 0.92  & 5.40 ± 0.38   \\
 &\textbf{Ours}     & \textbf{99.60 ± 0.22}        & \textbf{92.08 ± 0.63}  & \textbf{41.00 ± 1.06} \\
 
\hline
\hline
 \parbox[t]{4mm}{\multirow{7}{*}{\rotatebox[origin=c]{90}{\texttt{Lille2}}}}
 &\gls{onion}                          & 15.32 ± 0.60        & 11.48 ± 1.31  & 1.04 ± 0.32    \\
  &\gls{onion hist}~\cite{onionhist}    & 99.68 ± 0.24        & 98.84 ± 0.27  & 40.56 ± 0.97  \\
  &\gls{random walk}~\cite{xview}       & 97.88 ± 0.95        & 95.64 ± 1.08  & 49.32 ± 1.53  \\
 &\gls{gosg}~\cite{gosmatch}           & 63.16 ± 0.73        & 57.36 ± 0.88  & 7.44 ± 0.29   \\
 &\gls{gosv}~\cite{gosmatch}           & 94.48 ± 0.37        & 90.20 ± 0.66  & 28.84 ± 0.67  \\
 &\gls{pointnet}~\cite{qi2017pointnet} & 22.40 ± 1.39        & 20.64 ± 0.77  & 2.96 ± 0.34   \\
 &\gls{sgnet}~\cite{semantic_graph}    & 71.28 ± 0.84        & 61.48 ± 1.53  & 5.44 ± 0.59   \\
 &\textbf{Ours}     & \textbf{99.80 ± 0.13}       & \textbf{98.88 ± 0.20}  & \textbf{51.96 ± 0.71}  \\
 
\hline
\hline
\parbox[t]{4mm}{\multirow{7}{*}{\rotatebox[origin=c]{90}{\texttt{Paris}}}}
 &\gls{onion}       & 10.20 ± 0.52        & 4.20 ± 0.51   & 2.20 ± 0.38   \\
&\gls{onion hist}~\cite{onionhist}  & \textbf{96.60 ± 0.55}       & \textbf{80.12 ± 0.91}  & \textbf{47.24 ± 1.55}  \\
&\gls{random walk}~\cite{xview} & 91.32 ± 1.48        & 70.16 ± 2.60  & 46.96 ± 3.01  \\
 &\gls{gosg}~\cite{gosmatch}        & 74.08 ± 0.57        & 34.80 ± 0.68  & 10.48 ± 0.48  \\
 &\gls{gosv}~\cite{gosmatch}        & 87.04 ± 0.74        & 54.68 ± 0.56  & 31.72 ± 1.61  \\
 &\gls{pointnet}~\cite{qi2017pointnet}    & 12.00 ± 1.04        & 5.28 ± 0.45   & 3.04 ± 0.64   \\
 &\gls{sgnet}~\cite{semantic_graph}       & 48.48 ± 1.04        & 20.40 ± 0.83  & 9.28 ± 0.74   \\
 &\textbf{Ours}     & 93.24 ± 0.64  & 67.00 ± 0.92  & 41.12 ± 0.47  \\
 
\hline
\hline
\parbox[t]{4mm}{\multirow{7}{*}{\rotatebox[origin=c]{90}{\texttt{IQmulus}}}}
 &\gls{onion}       & 75.92 ± 0.68        & 30.20 ± 1.24  & 42.68 ± 1.79  \\
&\gls{onion hist}~\cite{onionhist}  & \textbf{99.40 ± 0.22}        & 69.68 ± 1.31  & 66.04 ± 1.38  \\
&\gls{random walk}~\cite{xview} & 95.68 ± 2.24        & 67.32 ± 2.37  & 64.00 ± 3.34  \\
 &\gls{gosg}~\cite{gosmatch}        & 8.08 ± 0.35         & 16.76 ± 0.74  & 1.12 ± 0.10   \\
 &\gls{gosv}~\cite{gosmatch}       & 31.36 ± 0.54        & 31.08 ± 0.24  & 13.32 ± 1.34  \\
 &\gls{pointnet}~\cite{qi2017pointnet}     & 43.60 ± 1.44        & 19.20 ± 0.88  & 30.00 ± 1.08  \\
 &\gls{sgnet}~\cite{semantic_graph}       & 25.84 ± 1.55        & 23.28 ± 1.11  & 3.32 ± 0.39   \\
 &\textbf{Ours}     & 98.08 ± 0.24        & \textbf{70.20 ± 1.12}  & \textbf{66.40 ± 0.67}  \\
\hline
\hline
\end{tabular}
\end{center}
\end{table}
\subsection{Global Localization Performance}
We sampled 500 query positions from each dataset and follow the pipeline described in Section~\ref{sec:method} to compute the estimated poses. Table~\ref{tab:test res} shows the localization performance of \gls{dptl} compared to the baselines on the Paris-Rue-Lille and IQmulus datasets.  On \texttt{Lille1} and \texttt{Lille2}, \gls{dptl} performs much better than the other descriptors in all cases. On \texttt{Paris}, \gls{dptl} shows a general performance among the top3 descriptors. On \texttt{IQmulus}, while not strictly outperforming all baselines, \gls{dptl} still shows the best overall performance and shows competitive results in all testing scenarios. In contrast, \gls{gosg}, \gls{onion} and \gls{pointnet} perform poorly in all cases.

On \texttt{Lille1}, in the \textit{Self-localization} scenario, \gls{random walk}, \gls{onion hist} and \gls{gosv} have the descriptiveness to identify the constellations. However, since these descriptors only take vague spatial relations into account instead of specific distances among objects in the constellation, their descriptiveness can be weak when the number of spatial relations decreases. This causes the significant decrease in localization accuracy in the \textit{Fewer Objects} scenario. In contrast, \gls{dptl} learns directly from the graph with specific distances between each pair of objects. The descriptiveness is still powerful even if the constellations contain fewer semantic objects because the distances between objects are actively used in the feature extraction. In the \textit{Added Noise} scenario, all descriptors' performances decrease. However, \gls{dptl} still performs the best thanks to being exposed to all kinds of noise during training.

Compared to \texttt{Lille1} and \texttt{Lille2}, \texttt{IQmulus} and \texttt{Paris} have much fewer objects and a shorter length, reducing the statistical strength of these experiements. Besides, for the IQumulus dataset, the object density reduces after removing the densely distributed ``bollard" objects. The performance of \gls{random walk}, \gls{onion hist} and \gls{dptl} is much lower than on the \texttt{Lille1} datasets in the \textit{Fewer objects} experiment, with greater similarity across the methods. This shows the influence of the object density and the distribution on the performance. The ambiguity may result from the self-similarity of the constellations or visual aliasing at different places. When the object distributions are similar to each other at different places, none of the tested methods are able to tell them apart and \gls{dptl} cannot extract more distinct descriptors from these constellations. Nevertheless, \gls{dptl} still shows robustness against noise and performs generally well among the top3 descriptors.

\subsection{Computational Efficiency}\label{sec:time analysis}

Our experiment includes two evaluation phases, i.e., computing descriptors, and matching them with geometric verification. We compare the average timing on all the datasets for all the testing scenarios on an Intel(R) Core(TM) i7-8750H CPU. The results are summarized in Table~\ref{tab:compute res}.

\gls{pointnet}, \gls{sgnet}, \gls{dptl} and \gls{gosg} take longer computing time since they are computed iteratively for each object. In contrast, \gls{onion}, \gls{onion hist} and \gls{random walk} are computed once altogether for all objects. \gls{pointnet} and \gls{sgnet} are faster thanks to batch computing. \gls{dptl} is still comparable with \gls{gosg} where batch computing is missing as well.

\gls{dptl} costs the shortest matching time thanks to its great distinctiveness.

\begin{table}[t]
\setlength\tabcolsep{3pt}
\centering
\caption{Computational Time Results. \texttt{Compute}, \texttt{Match}, and \texttt{Total} refer to average descriptor computing, matching and total time in~\si{\second}.}
\label{tab:compute res}
\begin{tabular}{@{}ccccccccc@{}}
\toprule
& \gls{onion} & \gls{onion hist} & \gls{random walk} & \gls{gosg} & \gls{gosv} & \gls{pointnet} & \gls{sgnet}  & \textbf{Ours} \\ 
\midrule
\texttt{Compute} & 0.014& 0.015& \textbf{0.005} & 0.138& 0.022& 0.006& 0.063& 0.243 \\
\texttt{Match} & 9.523& 3.415& 3.986& 10.307& 4.33& 8.594& 6.08& \textbf{2.63} \\
\texttt{Total} & 9.537& 3.430& 3.991& 10.445& 4.352& 8.600& 6.143 & \textbf{2.873} \\
\bottomrule
\end{tabular}
\end{table}

\subsection{Ablation Study}\label{sec: training ablation}

We explore how to combine geometric information and semantic information by setting up experiments on different input types for our model with the same backbones and attention layers. The input types are as follows:
\begin{itemize}[leftmargin=*]
    \item $(x, y, z)$: only contains geometric information, \textit{i.e.} the coordinates of the object. This serves as a baseline for the value of semantic information.
    \item $(x, y, z, C_{integer})$: besides geometric information, also contains semantic information encoded by a scalar integer $C_{integer} \in \{1, 2...N_{class}\}$.
    \item $(x, y, z, C_{onehot})$: encodes semantic information with a one-hot encoding, \textit{i.e.} $C_{onehot} \in \{0, 1\}^{N_{class} \times 1}$.
    \item $(x, y, z, C_{embed})$: encodes semantic information with a learnable embedding $C_{embed} \in \mathbb{R}^{N_{embed} \times 1}$. We use $N_{embed} = 3$ in our experiments.
\end{itemize}
We can conclude from Table~\ref{tab:train res} that the combination of geometric and semantic information is better than only using geometric information. Furthermore, a reasonable way to combine the geometric and semantic information is to first learn an embedding for the semantic class, and then concatenate it together with the geometric coordinates. We use this embedding-format model trained on the simulated training dataset to infer descriptors for global localization. As we choose an embedding that is much smaller than the one-hot encoding of the classes we save significant resources in the ResGCN backbone.

\begin{table}[t]
\small
\setlength\tabcolsep{3pt}
\centering
\caption{Training results with different model inputs.}
\label{tab:train res}
\begin{tabular}{@{}lcccc@{}}
\toprule
\multirow{2}{*}{Model Inputs}& \multicolumn{4}{c}{Metrics}\\
\cmidrule{2-5}
& train loss & val loss & train top5 & val top5  \\ 
\midrule
$(x, y, z)$ & 0.145  & 0.148  & 67.2\% & 81.7\%             \\
$(x, y, z, C_{integer})$ & 0.123  & 0.126  & 85.7\% & 94.4\%    \\
$(x, y, z, C_{onehot})$  & \textbf{0.0786} & 0.145 & 98.9\% & 90.0\% \\
$(x, y, z, C_{embed})$ & 0.0805 & \textbf{0.102} & \textbf{99.1\%} & \textbf{98.1\%}\\
\bottomrule
\end{tabular}
\end{table}
\section{Conclusion}
\label{sec:conclusion}
In this paper, we presented a \gls{gcn} architecture based on \glspl{dgcn} and the attention mechanism to learn a set of object constellation descriptors for global localization. Instead of relying on appearances to describe local features, our method leverages high-level semantic scene understanding and large-scale context, which enables localization in noisy scenes. This is important since the number of detectable semantic classes is typically limited, leading to self-similarity when only considering single objects in isolation. We built up pipelines to extract \gls{dptl} from real-world point cloud data, and compared its global localization performance with handcrafted and learned constellation descriptors. The models are trained on simulation data for easy learning and data efficiency, but show good performance with a matching accuracy measured by the Top5 ratio close to 100\% on simulation data, and a competitive localization success rate on real-world datasets.

\addtolength{\textheight}{0cm}   






\bibliographystyle{IEEEtran}
\bibliography{root}

\end{document}